\documentclass[10pt,twocolumn,letterpaper]{article}

\usepackage{cvpr}              

\usepackage{graphicx}
\usepackage{amsmath}
\usepackage{amssymb}
\usepackage{multirow}
\usepackage{booktabs}
\usepackage{adjustbox}
\usepackage{array}
\usepackage{colortbl}
\usepackage{float,wrapfig}
\usepackage{stfloats}
\usepackage{hhline}
\usepackage{bbding}
\usepackage{makecell}

\usepackage[pagebackref,breaklinks,colorlinks]{hyperref}

\usepackage[accsupp]{axessibility}
\usepackage[capitalize]{cleveref}
\crefname{section}{Sec.}{Secs.}
\Crefname{section}{Section}{Sections}
\Crefname{table}{Table}{Tables}
\crefname{table}{Tab.}{Tabs.}

\def\cvprPaperID{*****} 
\def\confName{CVPR}
\def\confYear{2022}

\begin{document}

\title{MVP: Robust Multi-View Practice for Driving Action Localization}

\author{
Jingjie~Shang\textsuperscript{1}\thanks{Equal contribution. \textsuperscript{\ddag}Contact person.}
\quad Kunchang Li\textsuperscript{2}\footnotemark[1]
\quad Kaibin Tian\textsuperscript{1}\footnotemark[1]
\quad Haisheng Su\textsuperscript{1\ddag} 
\quad Yangguang Li\textsuperscript{1} \\
}

\maketitle
\begin{abstract}
Distracted driving causes thousands of deaths per year,
and how to apply deep-learning methods to prevent these tragedies has become a crucial problem.
In Track3 of the 6th AI City Challenge,
researchers provide a high-quality video dataset with densely action annotations.
Due to the small data scale and unclear action boundary,
the dataset presents a unique challenge to precisely localize all the different actions and classify their categories.
In this paper,
we make good use of the multi-view synchronization among videos,
and conduct robust Multi-View Practice (MVP) for driving action localization.
To avoid overfitting,
we fine-tune SlowFast with Kinetics-700 pre-training as the feature extractor.
Then the features of different views are passed to ActionFormer to generate candidate action proposals.
For precisely localizing all the actions,
we design elaborate post-processing,
including model voting, threshold filtering and duplication removal.
The results show that our MVP is robust for driving action localization, 
which achieves 28.49\% F1-score in the Track3 test set.
\end{abstract}
\section{Introduction}

With the widespread application of deep learning in peoples' daily life \cite{resnet,maskrcnn,FPN,slowfast,uniformer2},
safety driving with computer vision attracts more and more attention.
According to the report of the National Highway Transportation and Safety Administration (NHTSA),
distracted driving causes about 920,000 total accidents per year,
wherein more than 3,000 people dead and about 280,000 injured. 
Unfortunately,
due to the lack of high-quality dataset,
it is difficult to conduct naturalistic driving studies, 
e.g.,
detecting the action of the driver in the traffic environment.

AI CITY Challenge\cite{Naphade17AIC17,Naphade18AIC18,Naphade19AIC19,Naphade20AIC20,Naphade21AIC21} has been held for 5 times. In Track3 of the 6th AI CITY Challenge,
a high-resolution video dataset about distracted driving is provided,
whose target is to detect the distracted behavior activities of drivers.
This task can be seen as temporal action localization (TAL) for long untrimmed video \cite{bsn,bmn},
which requires detecting temporal intervals that contain the distracting actions and then recognizing their categories.
Compared with pure action recognition \cite{tsn,slowfast,uniformer},
action localization is more challenging because of the unclear boundaries.
Different from image detection object boundaries are clearly defined,
there might not be a sensible definition of the start and end of action,
especially for the complex actions like gymnastics and long-term actions such as cycling.

Previous deep learning methods for TAL can be divided into two types:
two-stage TAL and one-stage TAL.
For two-stage approaches \cite{bsn,bmn,daps,tca,zhao2017cuhk},
they first generate candidate video segments as action proposals,
then further refine their temporal boundaries and recognize their categories.
As for one-stage methods \cite{ssn,gtan,pbrnet,afsd},
they tackle proposal and classification simultaneously,
i.e.,
localizing action without any proposals.
Though two-stage TAL can not be trained end-to-end as one-stage TAL,
it can generate more flexible and accurate proposals.
Thus we design our TAL in a two-stage style.

Compared with popular benchmark THUMOS \cite{thumos} and ActivityNet \cite{activitynet},
the dataset in Track3 presents new challenge for action localization.
On one hand,
the dataset is relatively small (totally of 60 videos) but densely annotated (18 different action instances per video),
while there are 826 and 19,994 videos for THUMOS and ActivityNet respectively,
and they contain fewer instances in one video.
On the other hand,
the official evaluation requires accurate boundary detection,
in which the predicted starts and ends should locate within one second of the truth.
More importantly,
all videos are synchronously recorded from three static cameras,
thus the action time is aligned among different views and the camera movement is omitted.
To summarize,
the ideal TAL needs to precisely localize all the different actions and recognize their corresponding categories.

In this paper,
we propose the simple yet effective Multi-View Practice (MVP) for driving action localization.
Our framework consists of the feature extractor, proposal generator and elaborate post-processing.
To avoid overfitting,
we fine-tune SlowFast-R101 \cite{slowfast} with Kinetics-700 pre-training as the feature extractor.
Then the features of different views are passed to ActionFormer \cite{actionformer},
which can generate candidate action proposals.
For precisely localizing all the actions,
we first ensemble multi-view models for accurate proposal classification,
then filter wrong proposals via pre-defined thresholds,
and finally remove the repeated actions.
The results show that our MVP is robust for driving action localization, 
which achieves 28.49\% F1-score in the Track3 test set of the 6th AI CITY Challenge.





\section{Dataset}

\begin{figure*}[tp]
    \begin{center}
        \includegraphics[width=0.98\textwidth]{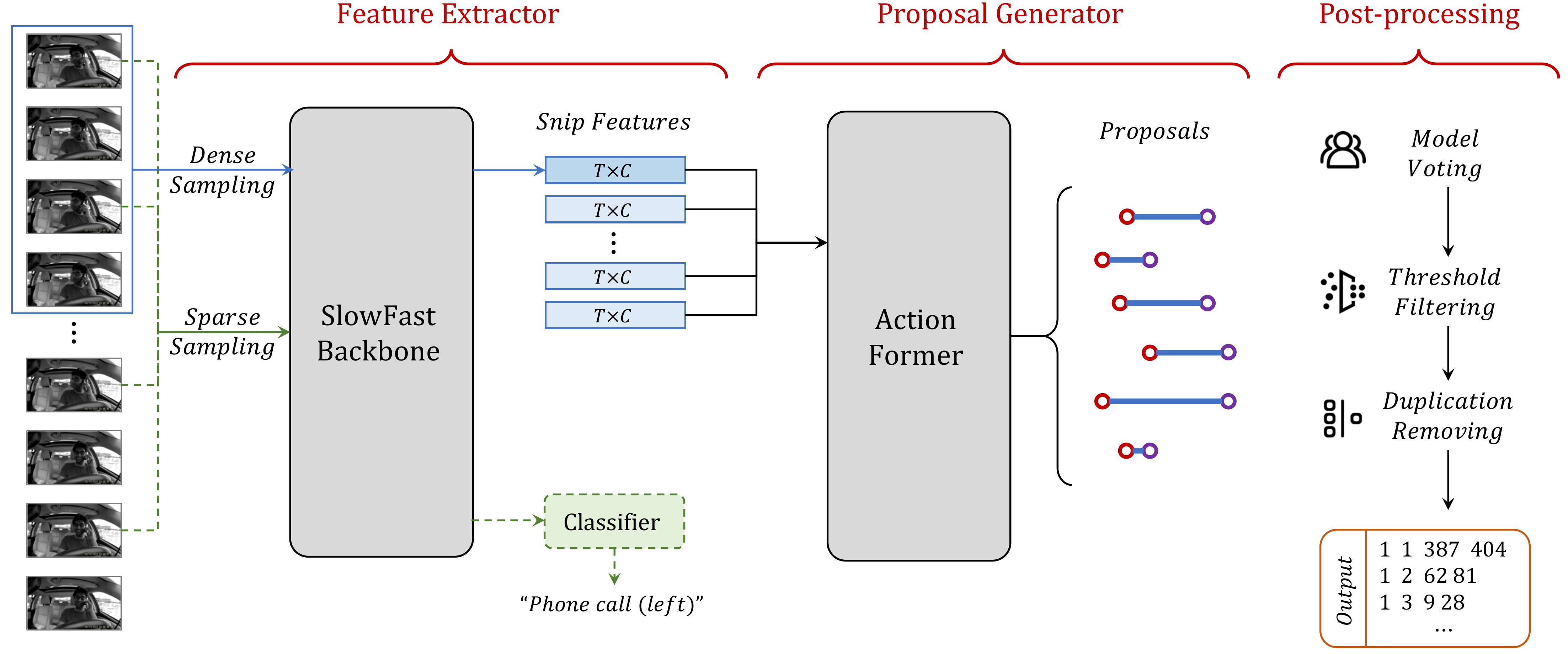}
    \end{center}
    \vspace{-0.6cm}
    \caption{
        \textbf{Overall architecture of our MVP framework.}
        Based on the synchronization among multi-view videos,
        we ensemble the models for better action recognition,
        and filter the redundant proposals for accurate boundaries detection. Here, T represents the length of snippet feature sequences and C represents the feature dimension.
    }
    \vspace{-0.3cm}
    \label{fig:framework}
\end{figure*}

We first analyze the video dataset SynDD1\cite{SynDD1} in Track3,
which help us conduct effective practices:
  \vspace{-1mm}
\begin{itemize}
  \item \textbf{Small but densely annotated}: The dataset only consists of 90 videos (totally of about 14 hours) from 15 volunteers,
  who pretend to do 18 different distracting actions (such as drinking, eating, yawning and singing) in a real vehicle cockpit. 
  Most of the durations of actions range from 10s to 30s, 
  and the interval durations between the two actions are usually around 10s.
  Besides, 
  some actions are confusing, 
  like "Talk to passenger at the right" and "Talk to passenger at backseat".
  \vspace{-2mm}
  \item \textbf{Multi-view synchronization}: For each driver,
  there are three static cameras to record his/her actions from different angles simultaneously (dashboard, rear and right).
  Moreover,
  each driver performed the distracted actions twice,
  with and without appearance blocks such as sunglasses and hat).
  \vspace{-2mm}
  \item \textbf{Need precise boundaries}: The official evaluation requires accurate boundary detection,
  in which the predicted starts and ends should locate within one second of the truth.
  All the different actions should be localized and recognized,
  while the repeated proposals are not accepted.
\end{itemize}
  \vspace{-1mm}
These three properties present unique challenges.

In our experiments,
the dataset is separated into three official sub-datasets,
A1 for training, 
and A2 and B for testing. 
Each sub-dataset contains 5 different drivers, 
with a total of 30 videos (3 views $\times$ 2 appearance blocks $\times$ 5 drivers). 
A1, as the training set, has the matching ground truth labels which are manually annotated. 
The labels include start time, 
end time, 
and action categories.
Based on A1, 
we divided it into A1-train and A1-val as our training dataset and validation dataset, 
which contained 4 drivers and 1 driver respectively. 

Accordingly, 
uncertain duration of actions, unclear boundaries and confusing classes all make temporal action localization a difficult task on this Track. 
We need to make good use of the multi-view videos and design a task-specific TAL framework to address these problems.



\section{Method}

\subsection{Overview}

The overall architecture of our Multi-View Practice (MVP) framework is shown in Figure \ref{fig:framework}.
Our MVP is composed of three main
components: (1) Feature Extractor, (2) Proposal Generator and (3) Post-processing. We explain each of the components in the following sections.
We first fine-tune the powerful SlowFast\cite{slowfast} with sparse sampling. 
Then the trained SlowFast can be used as the feature extractor via dense sampling. 
On the basis of snippet feature sequences, we use ActionFormer\cite{actionformer} to generate proposals.
Finally, we classify the proposals and conduct post-processing with multi-view knowledge.

\subsection{Feature Extractor}
\label{subsec:mvfe}
Before extracting features, 
we need to train video models for driving action recognition. 
The backbone in front of the fully-connected layer is taken as feature extractors. 
There are 5 drivers in A1 dataset, 
thus we split the videos of \textit{user\_id\_49381} as A1-val. 
The remaining videos will be divided into A1-train.
Then we split all the videos into action clips according to the time annotation,
wherein the background clips between two actions are ignored.

There are two mainstream structures for action recognition,
Convolutional Neural Networks (CNNs) \cite{resnet,c3d,slowfast} and Vision Transformers (ViTs)\cite{vit,viViT}.
Due to the inductive bias of CNNs, 
they are more powerful for small datasets,
while ViTs do not generalize well when trained on insufficient amounts of data \cite{vit}.
Since the dataset in Track3 only consists of 30 training videos,
we adopt SlowFast-R101\cite{slowfast} with Kinetics-700\cite{k700} pre-training as the backbone.
We choose the hyper-parameters via adequate adjustments.
Specifically,
we adopt the AdamW optimizer\cite{adamw} and set the weight decay penalty as 0.01.
In the warm-up stage, the learning rate starts at 1e-6 and increases to 5e-5 after 10 epochs.
Then, the model is trained for 190 epochs and the learning rate is decreased to 1e-6 by cosine annealing \cite{cosine}.
We use cross-entropy loss for training.
More importantly, 
to avoid overfitting,
we adopt the strong augmentations as in \cite{uniformer},
including mixup\cite{mixup}, cutmix\cite{cutmix}, Random Erasing \cite{erasing} and Rand Augment \cite{randaugment}.
We do not apply horizontal flip since some action categories are sensitive to it,
such as `Text (right)' and `Text (left)'.
Through exponential moving average \cite{ema}, we select the last checkpoint as the final model.
Note that when extracting features, 
the checkpoint is finally obtained by training all 5 drivers' videos in A1.

\begin{figure}[tp]
    \begin{center}
        \includegraphics[width=0.5\textwidth]{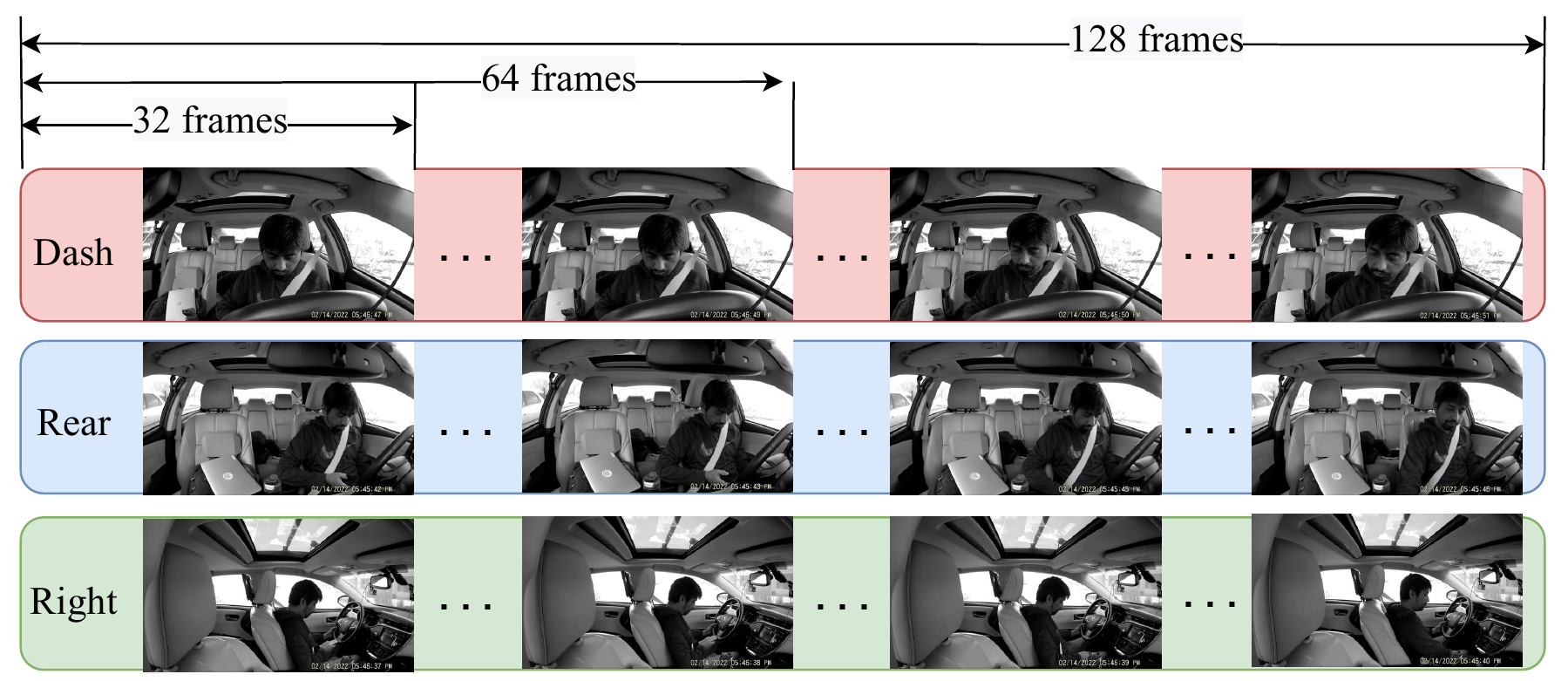}
    \end{center}
    \vspace{-0.6cm}
    \caption{
        \textbf{Sampling in three views.} Each row corresponds to one view. The range of driver action varies with the number of frames. We trained a total of 9 models with different views (dashboar/rear/right) and frame numbers (32/64/128).
    }
    \vspace{-0.3cm}
    \label{fig:feature_extract}
\end{figure}

Due to the different visual effects of views: dashboard, rear side and right side, 
we train the classification models for each view respectively. 
In each view, 
we also sample different numbers of frames. 
As shown in \ref{fig:feature_extract}, we train a total of 9 models with sparse sampling,
because the dense sampling is not effective enough as shown in Table \ref{tab:cls_setting}.
These multi-view models contribute to powerful action recognition.
When extracting snippet features,
we adopt dense sampling as in \cite{bsn} for accurate boundary detection.
Besides,
to generate robust temporal proposals,
we adopt four groups of features as shown in Table \ref{tab:proposal}.
The rear-side view features are abandoned because of the poor performance.

\subsection{Proposal Generator}
\label{subsec:pg}

In this section, we utilize snippet feature sequences of A1 to train the proposal generator, and generate proposals on A2. 
When training the proposal generator, snippet feature sequences have been fixed, which are detached from the feature extractor stage.
There are three reasons why we choose ActionFormer\cite{actionformer} as the core proposal generator: 
(1) Different from video modeling that contains a lot of redundant information,
snippet feature sequences already contain sufficient high-level semantic information, so we pay more attention to long-distance information interaction. 
(2) Transformer is suitable for time series data. Therefore, it is very favorable for mining time series context information and the interaction between them. 
(3) ActionFormer doesn't have the concept of anchor windows or sliding windows, which saves a lot of computing resources. At the same time, it draws lessons from the feature pyramid network and integrates multiple scales features, which can help the model cover action segments of different lengths.

We adopt the same training setting as \cite{actionformer},
the learning rate, warm-up epoch, total epoch and weight decay are set to 0.001, 5, 50 and 0.05 respectively.
In terms of model hype-parameters, we modified the longest sequence length as 2880 to adapt to the current data.
For the Non-Maximum Suppression(NMS)\cite{softnms},
we compare hard NMS and soft NMS and find that the soft NMS is better than hard NMS. 
On the basis of soft NMS, 
we also make different attempts for thresholds and get the best threshold of 0.1.
For different snippet feature sequences, 
we train different proposal generators respectively, so that the generated proposals can produce better fusion in the subsequent post-processing stage.
It is worth noting that we don't use the ActionFormer classifier because its classification performance is not ideal. 
The classifications for proposals are carried out in the next part.

\subsection{Post-processing}
Inspired by \cite{zheng2020vehiclenet}, we use a lot of post-processing to improve the performance.
We have trained multi-view driving action classification models in \ref{subsec:mvfe} and we have get proposals in \ref{subsec:pg}. Then, we get the driver action prediction results of the proposals through the classification models. The information from multiple views can complement each other, so each proposal is predicted by multi-view classification models and the prediction probabilities are averaged.

\begin{figure}[!tp]
    \begin{center}
        \includegraphics[width=0.35\textwidth]{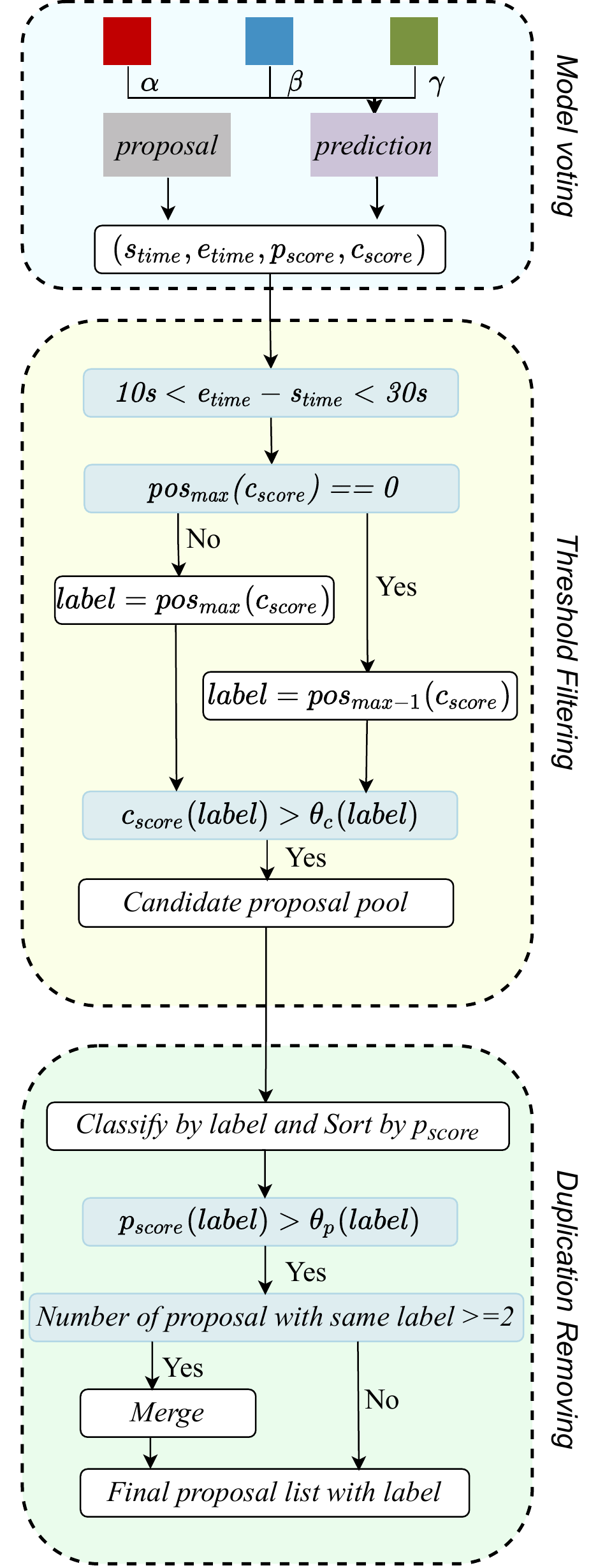}
    \end{center}
    \vspace{-0.6cm}
    \caption{
        \textbf{The flow diagram of post-processing} 
    }
    \vspace{-0.3cm}
    \label{fig:post}
\end{figure}

This section will introduce the post-processing for the results from our models. As depicted in Figure \ref{fig:post} post-processing can be mainly divided into three stages: Model voting, Threshold Filtering, and Duplication Removing. Our post-processing objects include proposals and corresponding predictions. The proposals come from the proposal generator, which contains start time($s_{time}$), end time($e_{time}$), and proposal confidence score ($p_{score}$). The number is $N$. In addition, for the predictions corresponding to the proposal, we used the classifier of the feature extractor to recognize the duration of each proposal and obtained the classification score vector. The vector contains the normalized score for each class. The size of each vector is $N\times18$.

For multiple models, we adopted the strategy of model voting to fuse the models. Specifically, each feature extraction model can be trained with data from three different views, and then three different classification score vectors can be obtained. We fuse the vectors from multiple views by weighted summation, and $a1,a2,a3$ represents the weight values respectively, which can be formulated as:
\begin{align} 
Pred = a1 \cdot P_{Dash}+a2 \cdot P_{Rear}+a3 \cdot P_{Right}
\end{align}
$P_{Dash},P_{Rear},P_{Right}$ represents the vector of different view respectively , and $Pred$ represents the classification score vector after fusion. In our typical experiment, $a1,a2,a3$ is set to 0.6,0.2,0.2. On this basis, we provide three different frame sampling rate setting models on the same feature extraction model. The number of sampling frames of each proposal is 32,64 and128. For the three sampling rates, we also use the weighted summation for fusion:
\begin{align}
c_{score} = \alpha \cdot Pred_{128}+\beta \cdot Pred_{64}+\gamma \cdot Pred_{32}
\end{align}
$c_{score}$ represents the classification score vector after final fusion. Typically, $\alpha,\beta,\gamma$ is set to 0.5,0.25,0.25. To summarise, for the same proposal list, a total of 9 classification score vectors are fused.

After the model voting is completed, the $c_{score}$ can be concatenated to the proposal information forming a complete prediction result with time and score. For this coarse result, we will further by filtering two scores. During the training, 5 drivers were alternately taken as validation dataset. And the classification scores and proposal confidence score of the correct results were counted on the validation dataset. We calculated the minimum values of the two scores for different classes, thus forming the threshold of classification ($\theta_{c}$) and threshold of proposal confidence($\theta_{p}$). According to our prior on the duration time, we first filtered the duration which is less than 10s or longer than 30s. And then we'll verify whether the predicted class is 0. Function $pos_{max}$ and $pos_{max-1}$ can return the positions corresponding to the max score and secondary score in the classification score vectors, namely label. If the class is 0, we will select the class with the second-highest classification score as the final class. At last, if the classification score is greater than the threshold of this class, it will be considered as a valid proposal and put into the candidate proposal pool.

According to the challenge stipulation, an action class only appears once in a video. Accordingly, in order to select the best proposal for a class, we divided the proposal according to the predicted class, and arranged it by confidence score in descending order. For sorted candidate pools, we first filtered the proposals whose confidence scores are lower than the threshold of the corresponding class. Then, if only one proposal remains in the one class, it will be included in the final result. If not, we will consider merging the two proposals with the highest and second-highest scores. If their start and end time differences are both less than 2s, both start time and end time will be averaged and two proposals will be merged into one proposal. If not, both proposals are retained. So far,  post-processing has been completed and gain the final proposal results.
\section{Experiments}

In this section, we show the experiment results.
Section \ref{subsec:eoar} reports the experiments on action recognition, Section \ref{subsec:eopg} reports the experiments on proposal generation and Section \ref{subsec:eoad} reports the results after post-processing. Finally, we show the leaderboard of Track3.



\subsection{Experiments on Action Recognition}
\label{subsec:eoar}

As shown in Table \ref{tab:backbone},
we compare the performance between powerful CNN (i.e., SlowFast-R101 \cite{slowfast}) and ViT (UniFormer-B \cite{uniformer}).
Both models are pre-trained with the large-scale datasets.
The results are tested with 3 crops by default.
It shows that the CNN-based model performs better for small datasets even with weak data augmentation (50.4\% vs. 21.3\%), where the weak augmentation only includes simple Rand Augment\cite{randaugment}.
When training with strong augmentation (see Section \ref{subsec:mvfe}),
both models improve the performance,
especially for the ViT-based model.
We adopt SlowFast-R101 for better action recognition.

In Table \ref{tab:cls_setting},
we further compared the performance of different settings.
The results clearly show that sparse sampling is better than dense sampling (83.3\% vs. 63.9\%).
Besides,
the single dashboard view is more powerful than the right-side and rear-side view,
but all the views are complementary,
thus ensembling different views can further improve the performance.
Finally,
training with more frames consistently increases the accuracy.

\begin{table}[tb]
\centering
\caption{The top-1 accuracy of different backbones. All the models are trained with 32 frames from the dashboard view.}
\vspace{-2mm}
\begin{tabular}{lcc|c}
\Xhline{1.0pt}
Model          & Pretrain & Augmentation & Acc \\ \hline
UniFormer-B     & K600    & weak  & 21.3   \\
SlowFast-R101     & K700    & weak & 50.4   \\ \hline
UniFormer-B     & K600    & strong  & 75.0   \\
\textbf{SlowFast-R101}     & \textbf{K700}    & \textbf{strong} & \textbf{83.3}   \\
\Xhline{1.0pt}
\end{tabular}
\label{tab:backbone}
\end{table}

\begin{table}[tb]
\centering
\caption{The top-1 accuracy of different sampling methods, views and frame numbers for SlowFast-R101.}
\vspace{-2mm}
\begin{tabular}{lcc|c}
\Xhline{1.0pt}
View          & Frame & Sampling & Acc \\ \hline
Dash     & 32    & Dense  & 63.9   \\
\textbf{Dash}     & \textbf{32}    & \textbf{Sparse} & \textbf{83.3}   \\ \hline
Right         & 32    & Sparse & 80.6   \\
Rear          & 32    & Sparse & 77.8   \\
Dash+Right & 32   & Sparse  & 91.7  \\ 
\textbf{Dash+Right+Rear} & \textbf{32}   & \textbf{Sparse}  & \textbf{92.5}  \\  \hline
Dash     & 64    & Sparse  & 85.2   \\
\textbf{Dash}     & \textbf{128}   & \textbf{Sparse}  & \textbf{87.1}  \\
\Xhline{1.0pt}
\end{tabular}
\label{tab:cls_setting}
\end{table}






\subsection{Experiments on Proposal Generation}
\label{subsec:eopg}

Table \ref{tab:proposal} shows the comparisons among different proposal generators.
It shows that the proposal generator trained by the right-side view achieves the best performance at @25, @50 and @100, 
but dashboard snippet feature sequences achieve the best AR at @150,
which demonstrates the complementarity between views.
Besides, 
in the same view, 
the number of frames and stride also affect the quality of temporal proposals.
To make good use of multi-view knowledge,
we try to integrate these proposals in post-processing.




\begin{table}[!tb]
\centering
\setlength\tabcolsep{2.5pt}
\caption{The AR@AN of different proposal generators. Following \cite{bsn}, we set the evaluation indicator  AR@AN to measure proposal generators' performance,
where AR means average recall and AN measns average number of proposals.}
\vspace{-2mm}
\begin{tabular}{ccc|cccc}
\Xhline{1.0pt}
\multicolumn{3}{c|}{Proposal Generator} & \multicolumn{4}{c}{AR@AN} \\ \hline
View & \begin{tabular}[c]{@{}c@{}}Snippet\\ Length\end{tabular} & \begin{tabular}[c]{@{}c@{}}Snippet\\ Stride\end{tabular} & @25 & @50 & @100 & @150 \\ \hline
Dashboard & 128 & 8 & 0.222 & 0.365 & 0.551 & 0.769 \\
Dashboard & 128 & 32 & 0.204 & 0.352 & 0.552 & \textbf{0.774} \\
Dashboard & 32 & 8 & 0.224 & 0.365 & 0.549 & 0.758 \\
Right & 128 & 32 & \textbf{0.231} & \textbf{0.382} & \textbf{0.568} & 0.770 \\ 
\Xhline{1.0pt}
\vspace{-3mm}
\label{tab:proposal}
\end{tabular}
\end{table}

\begin{table}[t]
	\centering
	\caption{Ablation study of each component of our post-processing. The best results are highlighted in bold.}
    \vspace{-2mm}
	\centering
	\begin{tabular}{ccc|cc}
	    \Xhline{1.0pt}
		Model&Threshold&Duplication&\multirow{2}{*}{mIoU}&\multirow{2}{*}{Acc}\\
		Voting&Filtering&Removing&&\\
		\hline
		$\times$&$\times$&$\times$&0.516&0.412\\
		\checkmark&$\times$&$\times$&0.667&0.741\\
		\checkmark&\checkmark&$\times$&0.775&0.893\\
		\checkmark&\checkmark&\checkmark&\textbf{0.833}&\textbf{0.918}\\
		\Xhline{1.0pt}
	\end{tabular}
	\centering
	\label{tab:post}
\end{table}


\subsection{Experiments on Post-processing}
\label{subsec:eoad}


In order to verify the effectiveness of the post-processing strategy, we conducted ablation studies as shown in Table \ref{tab:post}. 
To this end, we designed two evaluation metrics. The first one is temporal intersection over union (tIoU). 
The larger tIoU means a more accurate predicted start time and end time of the proposal. 
For all proposals, we calculated their mean tIoU (mIoU). 
Since only the proposals whose start and end are located within one second of the ground truth are considered the correct proposal, 
we defined the proposal with tIoU greater than 0.9 as a time-positive proposal. 
For the time-positive proposals, 
we further calculate their classification accuracy. 
And we find that these two metrics are positively correlated with Challenge metrics. 
The results show the effectiveness and robustness of our different post-processing strategies,
all of which play an important role in the final action localization.

\subsection{Track3 Leaderboard}
\label{subsec:t3l}

Table \ref{tab:leaderboard} shows the leaderboard of Track3 in the 6th AI City Challenge.
Our MVP framework achieves 28.49\% F1-score on the test dataset and ranks 8th. 
It is worth mentioning that our result is very close to the 5th, 6th and 7th places,
which verifies the effectiveness of our MVP.

\begin{table}[!tb]
\centering
\caption{F1 score for algorithms on testing data, which is evaluated on the evaluation system.}
\vspace{-2mm}
\begin{tabular}{lll}
\Xhline{1.0pt}
Team             & Rank & F1-Score \\ \hline
VTCC-UTVM        & 1    & 0.3492   \\
Stargazer        & 2    & 0.3295   \\
CybercoreAI      & 3    & 0.3248   \\
OPPilot          & 4    & 0.3154   \\
SIS Lab          & 5    & 0.2921   \\
BUPT-MCPRL2      & 6    & 0.2905   \\
Winter is Coming & 7    & 0.2902   \\
\textbf{HSNB}             & \textbf{8}    & \textbf{0.2849}   \\
VCA              & 9    & 0.2710   \\
Tahakom          & 10   & 0.2706   \\ 
\Xhline{1.0pt}
\end{tabular}
\label{tab:leaderboard}
\end{table}
\section{Conclusion}

In this paper,
we conduct robust Multi-View Practice (MVP) for driving action localization,
wherein we apply multi-view knowledge in feature extractor, proposal generator and post-processing.
Our simple yet effective MVP achieves 28.49\% F1-score in the Track3 test set in the 6th AICity Challenge.
However,
There is a big gap between our method and the top-1 resolution.
It is mainly because our proposal generator overfits the small training data and predicts imprecise action boundaries,
which will be further explored in the future.


{\small
\bibliographystyle{ieee_fullname}
\bibliography{egbib}

\begin{thebibliography}{10}\itemsep=-1pt

\bibitem{viViT}
A. Arnab, M. Dehghani, G. Heigold, Chen Sun, Mario Lucic, and C. Schmid.
\newblock Vivit: A video vision transformer.
\newblock {\em ArXiv}, abs/2103.15691, 2021.

\bibitem{softnms}
Navaneeth Bodla, Bharat Singh, Rama Chellappa, and Larry~S Davis.
\newblock Soft-nms--improving object detection with one line of code.
\newblock In {\em ICCV}, pages 5561--5569, 2017.

\bibitem{activitynet}
Fabian Caba~Heilbron, Victor Escorcia, Bernard Ghanem, and Juan Carlos~Niebles.
\newblock Activitynet: A large-scale video benchmark for human activity
  understanding.
\newblock In {\em CVPR}, pages 961--970, 2015.

\bibitem{k700}
Joao Carreira, Eric Noland, Chloe Hillier, and Andrew Zisserman.
\newblock A short note on the kinetics-700 human action dataset.
\newblock {\em ArXiv}, abs/1907.06987, 2019.

\bibitem{randaugment}
Ekin~D Cubuk, Barret Zoph, Jonathon Shlens, and Quoc~V Le.
\newblock Randaugment: Practical automated data augmentation with a reduced
  search space.
\newblock In {\em CVPR Workshop}, pages 702--703, 2020.

\bibitem{vit}
Alexey Dosovitskiy, Lucas Beyer, Alexander Kolesnikov, Dirk Weissenborn,
  Xiaohua Zhai, Thomas Unterthiner, Mostafa Dehghani, Matthias Minderer, Georg
  Heigold, Sylvain Gelly, et~al.
\newblock An image is worth 16x16 words: Transformers for image recognition at
  scale.
\newblock {\em ArXiv}, abs/2010.11929, 2020.

\bibitem{daps}
Victor Escorcia, Fabian Caba~Heilbron, Juan~Carlos Niebles, and Bernard Ghanem.
\newblock Daps: Deep action proposals for action understanding.
\newblock In {\em ECCV}, pages 768--784, 2016.

\bibitem{slowfast}
Christoph Feichtenhofer, Haoqi Fan, Jitendra Malik, and Kaiming He.
\newblock Slowfast networks for video recognition.
\newblock In {\em ICCV}, pages 6202--6211, 2019.

\bibitem{maskrcnn}
Kaiming He, Georgia Gkioxari, Piotr Doll{\'a}r, and Ross Girshick.
\newblock Mask r-cnn.
\newblock In {\em ICCV}, pages 2961--2969, 2017.

\bibitem{resnet}
Kaiming He, Xiangyu Zhang, Shaoqing Ren, and Jian Sun.
\newblock Deep residual learning for image recognition.
\newblock In {\em CVPR}, pages 770--778, 2016.

\bibitem{thumos}
Haroon Idrees, Amir~R Zamir, Yu-Gang Jiang, Alex Gorban, Ivan Laptev, Rahul
  Sukthankar, and Mubarak Shah.
\newblock The thumos challenge on action recognition for videos “in the
  wild”.
\newblock {\em Computer Vision and Image Understanding}, 155:1--23, 2017.

\bibitem{FPN}
Alexander Kirillov, Ross Girshick, Kaiming He, and Piotr Doll{\'a}r.
\newblock Panoptic feature pyramid networks.
\newblock In {\em CVPR}, pages 6399--6408, 2019.

\bibitem{uniformer}
Kunchang Li, Yali Wang, Gao Peng, Guanglu Song, Yu Liu, Hongsheng Li, and Yu
  Qiao.
\newblock Uniformer: Unified transformer for efficient spatial-temporal
  representation learning.
\newblock In {\em ICLR}, 2022.

\bibitem{uniformer2}
Kunchang Li, Yali Wang, Junhao Zhang, Peng Gao, Guanglu Song, Yu Liu, Hongsheng
  Li, and Yu Qiao.
\newblock Uniformer: Unifying convolution and self-attention for visual
  recognition.
\newblock {\em ArXiv}, abs/2201.09450, 2022.

\bibitem{afsd}
Chuming Lin, Chengming Xu, Donghao Luo, Yabiao Wang, Ying Tai, Chengjie Wang,
  Jilin Li, Feiyue Huang, and Yanwei Fu.
\newblock Learning salient boundary feature for anchor-free temporal action
  localization.
\newblock In {\em CVPR}, pages 3320--3329, 2021.

\bibitem{bmn}
Tianwei Lin, Xiao Liu, Xin Li, Errui Ding, and Shilei Wen.
\newblock Bmn: Boundary-matching network for temporal action proposal
  generation.
\newblock In {\em ICCV}, pages 3889--3898, 2019.

\bibitem{bsn}
Tianwei Lin, Xu Zhao, Haisheng Su, Chongjing Wang, and Ming Yang.
\newblock Bsn: Boundary sensitive network for temporal action proposal
  generation.
\newblock In {\em ECCV}, pages 3--19, 2018.

\bibitem{pbrnet}
Qinying Liu and Zilei Wang.
\newblock Progressive boundary refinement network for temporal action
  detection.
\newblock In {\em AAAI}, pages 11612--11619, 2020.

\bibitem{gtan}
Fuchen Long, Ting Yao, Zhaofan Qiu, Xinmei Tian, Jiebo Luo, and Tao Mei.
\newblock Gaussian temporal awareness networks for action localization.
\newblock In {\em CVPR}, pages 344--353, 2019.

\bibitem{cosine}
Ilya Loshchilov and Frank Hutter.
\newblock Sgdr: Stochastic gradient descent with warm restarts.
\newblock {\em ArXiv}, abs/1608.03983, 2016.

\bibitem{adamw}
Ilya Loshchilov and Frank Hutter.
\newblock Decoupled weight decay regularization.
\newblock In {\em ICLR}, 2019.

\bibitem{Naphade17AIC17}
Milind Naphade, David~C. Anastasiu, Anuj Sharma, Vamsi Jagrlamudi, Hyeran Jeon,
  Kaikai Liu, Ming-Ching Chang, Siwei Lyu, and Zeyu Gao.
\newblock The nvidia ai city challenge.
\newblock In {\em CVPR Workshop}, Santa Clara, CA, USA, 2017.

\bibitem{Naphade18AIC18}
Milind Naphade, Ming-Ching Chang, Anuj Sharma, David~C. Anastasiu, Vamsi
  Jagarlamudi, Pranamesh Chakraborty, Tingting Huang, Shuo Wang, Ming-Yu Liu,
  Rama Chellappa, Jenq-Neng Hwang, and Siwei Lyu.
\newblock The 2018 nvidia ai city challenge.
\newblock In {\em CVPR Workshop}, pages 53--–60, 2018.

\bibitem{Naphade19AIC19}
Milind Naphade, Zheng Tang, Ming-Ching Chang, David~C. Anastasiu, Anuj Sharma,
  Rama Chellappa, Shuo Wang, Pranamesh Chakraborty, Tingting Huang, Jenq-Neng
  Hwang, and Siwei Lyu.
\newblock The 2019 ai city challenge.
\newblock In {\em CVPR Workshop}, page 452–460, June 2019.

\bibitem{Naphade21AIC21}
Milind Naphade, Shuo Wang, David~C. Anastasiu, Zheng Tang, Ming-Ching Chang,
  Xiaodong Yang, Yue Yao, Liang Zheng, Pranamesh Chakraborty, Christian~E.
  Lopez, Anuj Sharma, Qi Feng, Vitaly Ablavsky, and Stan Sclaroff.
\newblock The 5th ai city challenge.
\newblock In {\em CVPR Workshop}, June 2021.

\bibitem{Naphade20AIC20}
Milind Naphade, Shuo Wang, David~C. Anastasiu, Zheng Tang, Ming-Ching Chang,
  Xiaodong Yang, Liang Zheng, Anuj Sharma, Rama Chellappa, and Pranamesh
  Chakraborty.
\newblock The 4th ai city challenge.
\newblock In {\em CVPR Workshop}, page 2665–2674, June 2020.

\bibitem{ema}
Boris~T Polyak and Anatoli~B Juditsky.
\newblock Acceleration of stochastic approximation by averaging.
\newblock {\em SIAM journal on control and optimization}, 30(4):838--855, 1992.

\bibitem{tca}
Zhiwu Qing, Haisheng Su, Weihao Gan, Dongliang Wang, Wei Wu, Xiang Wang, Yu
  Qiao, Junjie Yan, Changxin Gao, and Nong Sang.
\newblock Temporal context aggregation network for temporal action proposal
  refinement.
\newblock In {\em CVPR}, pages 485--494, 2021.

\bibitem{SynDD1}
Mohammed~Shaiqur Rahman, Archana Venkatachalapathy, Anuj Sharma, Jiyang Wang,
  Senem~Velipasalar Gursoy, David Anastasiu, and Shuo Wang.
\newblock Synthetic distracted driving (syndd1) dataset for analyzing
  distracted behaviors and various gaze zones of a driver, 2022.

\bibitem{c3d}
Du Tran, Lubomir Bourdev, Rob Fergus, Lorenzo Torresani, and Manohar Paluri.
\newblock Learning spatiotemporal features with 3d convolutional networks.
\newblock In {\em ICCV}, pages 4489--4497, 2015.

\bibitem{tsn}
L. Wang, Yuanjun Xiong, Zhe Wang, Y. Qiao, D. Lin, X. Tang, and L. Gool.
\newblock Temporal segment networks: Towards good practices for deep action
  recognition.
\newblock In {\em ECCV}, 2016.

\bibitem{cutmix}
Sangdoo Yun, Dongyoon Han, Seong~Joon Oh, Sanghyuk Chun, Junsuk Choe, and
  Youngjoon Yoo.
\newblock Cutmix: Regularization strategy to train strong classifiers with
  localizable features.
\newblock In {\em ICCV}, pages 6023--6032, 2019.

\bibitem{actionformer}
Chenlin Zhang, Jianxin Wu, and Yin Li.
\newblock Actionformer: Localizing moments of actions with transformers.
\newblock {\em ArXiv}, abs/2202.07925, 2022.

\bibitem{mixup}
Hongyi Zhang, Moustapha Cisse, Yann~N. Dauphin, and David Lopez-Paz.
\newblock mixup: Beyond empirical risk minimization.
\newblock In {\em ICLR}, 2018.

\bibitem{ssn}
Yue Zhao, Yuanjun Xiong, Limin Wang, Zhirong Wu, Xiaoou Tang, and Dahua Lin.
\newblock Temporal action detection with structured segment networks.
\newblock In {\em ICCV}, pages 2914--2923, 2017.

\bibitem{zhao2017cuhk}
Yue Zhao, Bowen Zhang, Zhirong Wu, Shuo Yang, Lei Zhou, Sijie Yan, Limin Wang,
  Yuanjun Xiong, D Lin, Y Qiao, et~al.
\newblock Cuhk \& ethz \& siat submission to activitynet challenge 2017.
\newblock {\em ArXiv}, abs/1710.08011, 2017.

\bibitem{zheng2020vehiclenet}
Zhedong Zheng, Tao Ruan, Yunchao Wei, Yi Yang, and Tao Mei.
\newblock Vehiclenet: Learning robust visual representation for vehicle
  re-identification.
\newblock {\em IEEE Transactions on Multimedia}, 23:2683--2693, 2020.

\bibitem{erasing}
Zhun Zhong, Liang Zheng, Guoliang Kang, Shaozi Li, and Yi Yang.
\newblock Random erasing data augmentation.
\newblock In {\em AAAI}, pages 13001--13008, 2020.

\end{thebibliography}
}

\end{document}